%% file: PortraitEditor.tex
\documentclass[acmtog]{acmart}

\AtBeginDocument{%
  }

\copyrightyear{2024}
\acmYear{2024}
\setcopyright{acmlicensed}\acmConference[SA Conference Papers '24]{SIGGRAPH Asia 2024 Conference Papers}{December 3--6, 2024}{Tokyo, Japan}
\acmBooktitle{SIGGRAPH Asia 2024 Conference Papers (SA Conference Papers '24), December 3--6, 2024, Tokyo, Japan}
\acmDOI{10.1145/3680528.3687601}
\acmISBN{979-8-4007-1131-2/24/12}

\acmSubmissionID{411}

\newcommand{\revision}[1]{{#1}}

\begin{document}

\title{Portrait Video Editing Empowered by Multimodal Generative Priors}
\author{Xuan Gao}
\email{gx2017@mail.ustc.edu.cn}
\affiliation{%
	\institution{University of Science and Technology of China}
	\country{China}
}

\author{Haiyao Xiao}
\email{xhy1999512@mail.ustc.edu.cn}
\affiliation{%
	\institution{University of Science and Technology of China}
	\country{China}
}

\author{Chenglai Zhong}
\email{zcl2017@mail.ustc.edu.cn}
\affiliation{%
	\institution{University of Science and Technology of China}
	\country{China}
}

\author{Shimin Hu}
\email{sa23001018@mail.ustc.edu.cn}
\affiliation{%
	\institution{University of Science and Technology of China}
	\country{China}
}

\author{Yudong Guo}
\email{yudong@ustc.edu.cn}
\affiliation{%
	\institution{University of Science and Technology of China}
	\country{China}
}

\author{Juyong Zhang}
\email{juyong@ustc.edu.cn}
\authornote{Corresponding author (\href{mailto:juyong@ustc.edu.cn}{juyong@ustc.edu.cn}).}
\affiliation{%
	\institution{University of Science and Technology of China}
	\country{China}
}

\renewcommand{\shortauthors}{Gao et al.}

\begin{abstract}


We introduce PortraitGen, a powerful portrait video editing method that achieves consistent and expressive stylization with multimodal prompts. Traditional portrait video editing methods often struggle with 3D and temporal consistency, and typically lack in rendering quality and efficiency. To address these issues, we lift the portrait video frames to a unified dynamic 3D Gaussian field, which ensures structural and temporal coherence across frames. Furthermore, we design a novel Neural Gaussian Texture mechanism that not only enables sophisticated style editing but also achieves rendering speed over 100FPS. Our approach incorporates multimodal inputs through knowledge distilled from large-scale 2D generative models. Our system also incorporates expression similarity guidance and a face-aware portrait editing module, effectively mitigating degradation issues associated with iterative dataset updates. Extensive experiments demonstrate the temporal consistency, editing efficiency, and superior rendering quality of our method. The broad applicability of the proposed approach is demonstrated through various applications, including text-driven editing, image-driven editing, and relighting, highlighting its great potential to advance the field of video editing. Demo videos and released code are provided in our project page: https://ustc3dv.github.io/PortraitGen/

\end{abstract}



\begin{teaserfigure}
\includegraphics[width=1\textwidth]{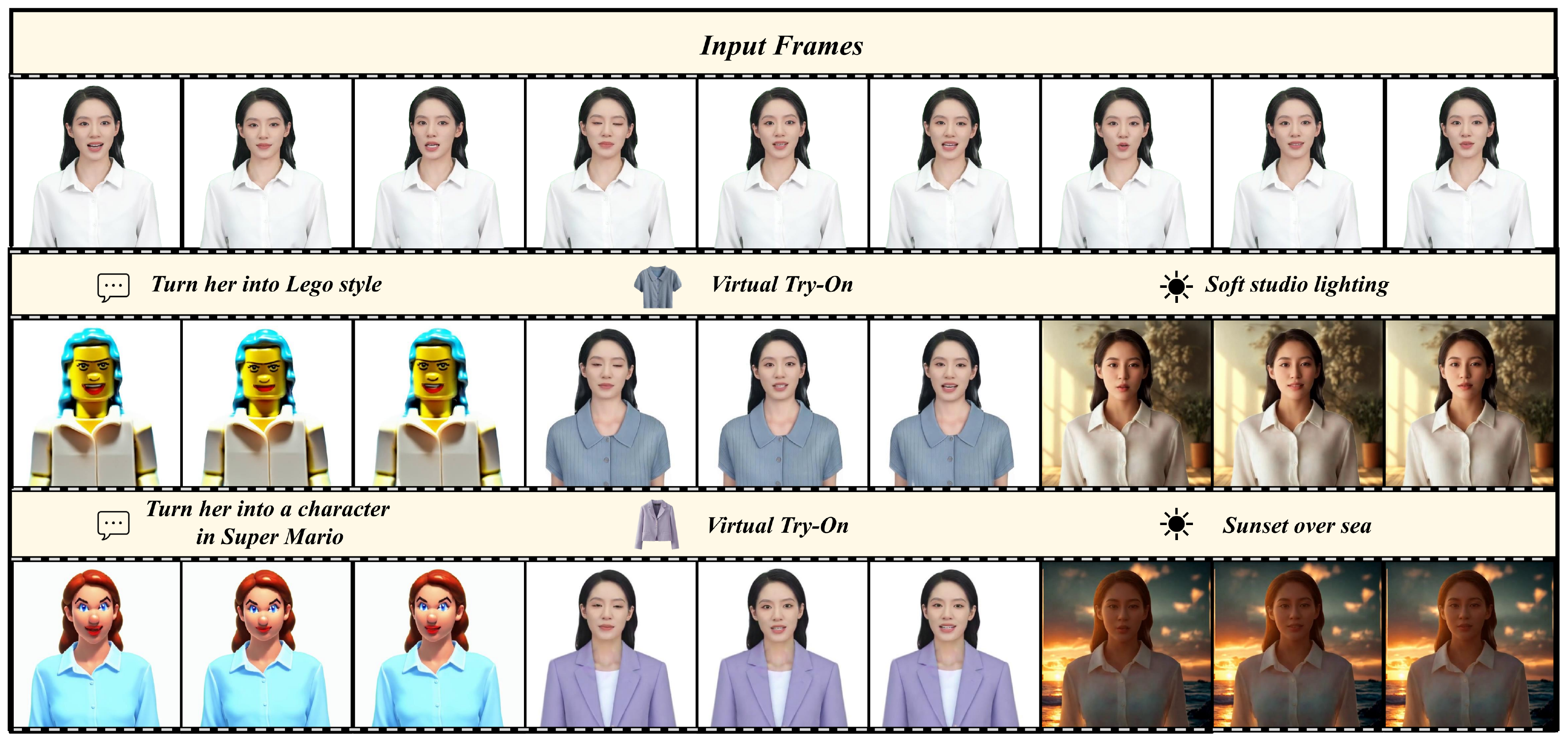}
\caption{PortraitGen is a powerful portrait video editing method that achieves consistent and expressive stylization with multimodal prompts. Given a monocular RGB video, our model could perform high-quality text driven editing, image driven editing and relighting.}
\end{teaserfigure}

\maketitle

\input{Sec01_Introduction.tex}

\input{Sec02_RelatedWork}

\input{Sec03_Method}

\input{Sec04_Experiments}

\input{Sec05_Ablation}

\input{Sec06_Conclusion}

\bibliographystyle{ACM-Reference-Format}
\bibliography{sample-base}






\end{document}

%% file: Sec01_Introduction.tex
\section{Introduction}
Portrait video editing has extensive applications in fields such as film, art, and AR/VR. Ensuring structural similarity and temporal consistency across the whole sequence, while enabling various functionalities and modalities, and achieving high-quality editing results, have always been challenging.

2D portrait editing has been studied a lot. Early works~\cite{yang2022Dual,yang2022Vtoonify,liu20223dfm} mainly adopt Generative Adversarial Network (GAN)~\cite{goodfellow2014generative} for editing or stylized animation based on style labels or reference images. By minimizing CLIP~\cite{clip} similarity, some works~\cite{patashnik2021styleclip,gal2021stylegannada,xia2021tedigan} successfully generate images based on text descriptions. However, this kind of works is limited by the representation ability of the GAN model. Recently, diffusion models~\cite{ho2020denoising} have shown great generation ability compared with GAN. Based on the denoising diffusion scheme, a lot of generative models, adapters, and finetuning methods are proposed to generate high-quality stylized portrait images. However, when editing portrait videos, these methods struggle to maintain temporal consistency across frames. 

To improve the continuity of edited video, some works choose to explore training-free video editing with pre-trained image diffusion models. They use dense correspondence, DDIM inversion~\cite{song2020denoising}, ControlNet~\cite{zhang2023adding}, or cross-frame attention to make editing aware of the motion or underlying structures of the original video. Other works turn to connect the frames in temporal dimension and train temporal attention to ensure temporal or multi-view consistency~\cite{guo2023animatediff,qin2023instructvid2vid}. However, due to the lack of 3D understanding and facial/body priors, they might fail to generate video results that is satisfying in quality and temporal consistency. Meanwhile, these methods need minutes of computation to generate only 1-second video clip due to the progressive sampling and complicated computation of the denoising process.

In this paper, we propose a portrait video editing system that is: (1) preserving portrait structure, (2) temporally consistent, (3) efficient, and (4) capable of multimodal editing requirements. Unlike previous works that focus solely on the 2D domain, we lift the portrait video editing problem into 3D to ensure 3D awareness. Additionally, we distill the multimodal editing knowledge from existing 2D generative models to facilitate high-quality editing.

\begin{figure}[t]
  \centering
  \includegraphics[width=0.97\linewidth]{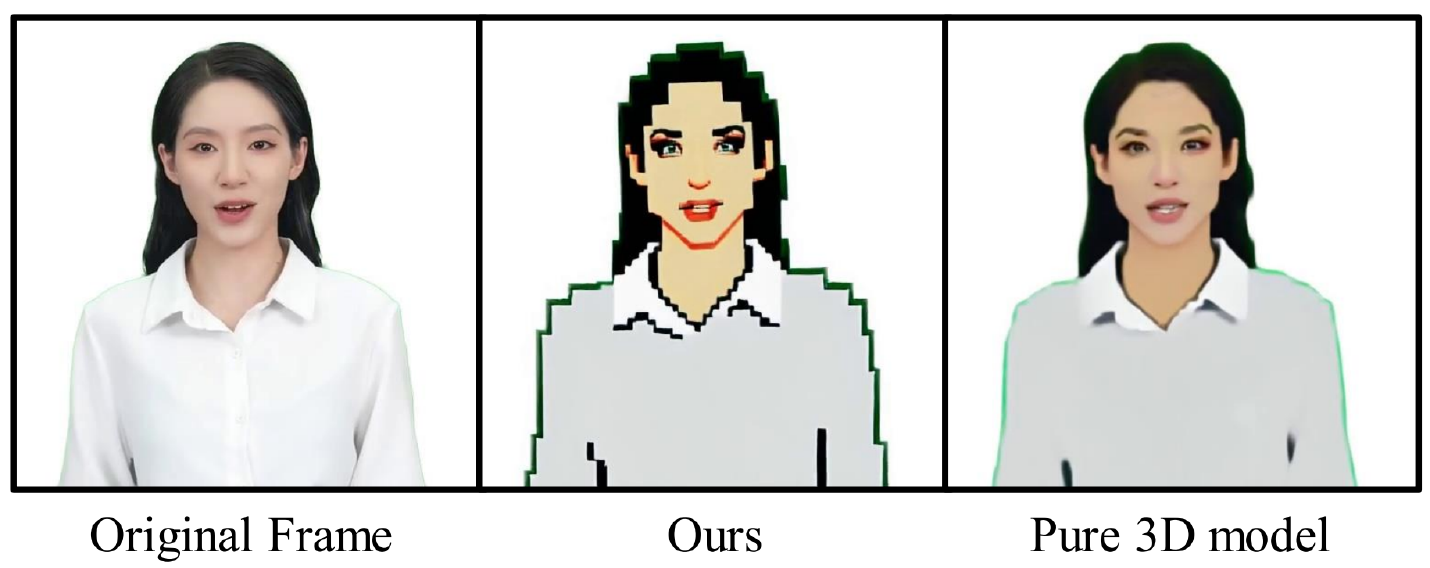}
  \caption{A totally 3D consistent model may not be an ideal solution for some styles. Many styles include intricate brush strokes and contour lines, which are actually not 3D consistent. Given the instruction `Turn her into pixel style', our edited portrait could exhibit pixel contour lines, which is crucial for this kind of stylization.}
  \label{fig:NGT_demonstrate}
  \vspace*{-5mm}
\end{figure}

Specifically, we employ 3D Gaussian Splatting (3DGS)~\cite{kerbl3Dgaussians} for consistent and efficient rendering. We embed the 3D Gaussian field on the surface of SMPL-X~\cite{SMPL-X:2019} to ensure structural and temporal consistency. Previous 3DGS-based portrait representations~\cite{xiang2024flashavatar,qian2023gaussianavatars} store spherical harmonic (SH) coefficients for each Gaussian and supervise the splatted image directly. However, although these kinds of representations may exhibit high-fidelity in the reconstruction task, they are not qualified for editing tasks. The reason behind this is that many styles include intricate brush strokes and contour lines, which are actually not totally 3D consistent. 
Directly fitting such signals with 3DGS can result in blurring or artifacts. Moreover, in some artistic styles, portraits often deviate greatly from real people, which calls for more expressive representations. Inspired by Neural Texture~\cite{thies2019deferred} and screen post-processing effects in non-photorealistic rendering, we store a learnable feature for each Gaussian instead of storing SH coefficients. We then employ a 2D neural renderer to transform the splatted feature map into RGB signals. This approach provides a more informative feature than SH coefficients and allows for a better fusion of splatted features, facilitating the editing of more complex styles. As demonstrated in Fig.~\ref{fig:NGT_demonstrate}, with the help of this Neural Gaussian Texture mechanism, our method supports editing whose styles are not completely 3D consistent and achieves rendering speed over 100FPS.

To distill the knowledge of 2D multimodal generative models into portrait video editing, we alternate between editing the dataset of video frames and updating the underlying 3D portrait, inspired by~\cite{instructnerf2023}. However, we find that naively using this iterative dataset update strategy may accumulate errors in expressions and facial structures, causing blurring and expression degradation. To address these issues, we design an expression similarity guidance term to ensure expression correctness. Additionally, we propose a face-aware portrait editing module to preserve facial structures. Experiments demonstrate that our scheme could effectively preserve personalized structures of original portrait videos and outperform previous works in quality, efficiency, and temporal consistency. Applications such as text-driven editing, image-driven editing, and relighting further underscore the effectiveness and multimodal generalizability of our approach.

In summary, the main contributions of our work include:
\begin{itemize}
\item We present PortraitGen, an expressive and consistent portrait video editing system. By lifting the 2D portrait video editing problem into 3D and introducing 3D human priors, it effectively ensures both 3D consistency and temporal consistency of the edited video.  
\item Our Neural Gaussian Texture mechanism enables richer 3D information and improves the rendering quality of edited portraits, and it helps to support complex styles.
\item Our expression similarity guidance and face-aware portrait editing module can effectively handle the degradation problems of iterative dataset update, and further enhance expression quality and preserve personalized facial structures.
\end{itemize}

%% file: Sec02_RelatedWork.tex
\section{Related Work}
\paragraph{Digital Portrait Representation}
Digital portrait representation has been studied for a long time. Blanz and Vetter proposed 3DMM~\cite{blanz1999morphable} to embed 3D head shape into several low-dimensional PCA spaces. The explicit head model has been further studied by a lot of following works. To improve its representation ability, some work extends it to multilinear models~\cite{cao2013facewarehouse,vlasic2006face}, and non-linear models~\cite{tran2018nonlinear,guo20213d}, articulated models~\cite{li2017learning}. They have been used for many applications. However, due to the limited representation ability, they fail to synthesize photo-realistic results. 

Implicit representations have been widely used in 3D modeling~\cite{wang2021neus,Song2024City,verbin2022ref} and editing~\cite{zhang2022arf,qiu2024deformable,neumesh}. They use neural functions to fit the radiance field, signed distance field, or occupancy field. A series of generative head models have been proposed~\cite{Chan2021pi,gu2021stylenerf,GIRAFFE,chan2022efficient,deng2022gram,or2022stylesdf,wang2023rodin}. Some works proposed parametric implicit head model~\cite{hong2021headnerf,zhuang2022mofanerf} or integrate 3D generative model with face priors~\cite{sun2023next3d,yue2023aniportraitgan,yue2022anifacegan} to realize animation. Although implicit representations could achieve satisfied rendering quality, they suffer from limited rendering efficiency.

Recently, 3D Gaussian Splatting (3DGS)~\cite{kerbl3Dgaussians} has been applied to digital head modeling. Because of its flexible representation and fast differentiable rasterizer, this kind of head model achieves remarkable performance in efficiency~\cite{xiang2024flashavatar,dhamo2023headgas} and fidelity~\cite{qian2023gaussianavatars,wang2024gaussianhead,xu2023gaussian}. There is also work adopting 3DGS for hair modeling and rendering~\cite{luo2024gaussianhair}.

\paragraph{Diffusion Model in Vision}
Denoising Diffusion Model~\cite{ho2020denoising} has showcased great generative ability in vision. Recent works can be classified into 2D image synthesis~\cite{brooks2022instructpix2pix,ruiz2023dreambooth,rombach2022highresolution,zhang2023adding} and 3D scene generation~\cite{instructnerf2023,poole2022dreamfusion,liu2024genn2n,GaussianEditor,chen2023gaussianeditor,tang2023dreamgaussian}. While these approaches can generate high-quality results from arbitrary text prompts, they mainly concentrate on generating or editing individual, static tasks and are not intended to directly edit dynamic scenes, especially 2D/3D portrait videos with complex motion. 

As a result, some researchers have shifted their focus to video tasks. The main challenge is the consistency between different frames. \revision{To solve this problem, some methods~\cite{wu2023tuneavideo,qi2023fatezero,wang2023zeroshot,tokenflow2023,ku2024anyv2v,molad2023dreamix,zhang2024towards} modify the latent space of the diffusion model and introduce cross-frame attention maps to enhance the consistency of the generated results.} However, purely modifying in attention space could not enable the model consistent in details. Rerender-A-Video~\cite{yang2023rerender} and CoDeF~\cite{ouyang2023codef} use optical flow to enhance fine-detailed consistency, which suffers from limited optical flow accuracy and struggles to model complex motion.

\paragraph{Portrait Editing}
The editing of the appearance and semantic attributes of digital humans has always attracted a lot of attention. Following the success of StyleGAN2~\cite{karras2020styleGAN2}, \revision{many researchers utilize pre-trained GAN model for facial editing or animation~\cite{Abdal_2021_styleflow,liu20223dfm,yang2022Dual,yang2022Vtoonify,kwon2022clipstyler,patashnik2021styleclip,tzaban2022stitch}.} However, due to limited generation ability of StyleGAN2, these methods fail to get robust results in complex motion.

To address this issue, researchers utilize 3D representations as geometric proxies to enhance the 3D consistency in editing. Some methods~\cite{canfes2022textavatar,aneja2023clipface} directly employ 3DMM (3D Morphable Model) as geometric representation and utilize generative models to generate corresponding UV textures. These methods suffer from the limited representation ability of mesh models and may lack personality in appearance and motion. \revision{Recent works~\cite{sun2022fenerf,sun2023next3d,bao2024geneavatar,abdal20233davatargan} utilize NeRF for the purpose of editing}, which is not efficient enough for many applications.

Many recent works use diffusion models to perform editing or generation tasks. Among them, 2D image works mainly focus on the generation and editing of face portraits~\cite{paraperas2024arc2face,tian2024emo}. With the help of Score Distillation Sampling~\cite{poole2022dreamfusion}, researchers tend to construct 3D avatars according to text prompt~\cite{han2023headsculpt,zhang2023dreamface}. For 3D avatar editing, Avatarstudio~\cite{mendiratta2023avatarstudio} proposes a view-and-time-aware Score Distillation Sampling to enable high-quality personalized editing across the view and time domain. Control4D~\cite{shao2023control4d} uses a generative adversarial strategy to handle inconsistency between different frames. Both methods require multi-view dynamic video sequences as input for avatar modeling, which are difficult to obtain for practical use.

%% file: Sec03_Method.tex
\begin{figure*}[t]
  \centering
  \includegraphics[width=\linewidth]{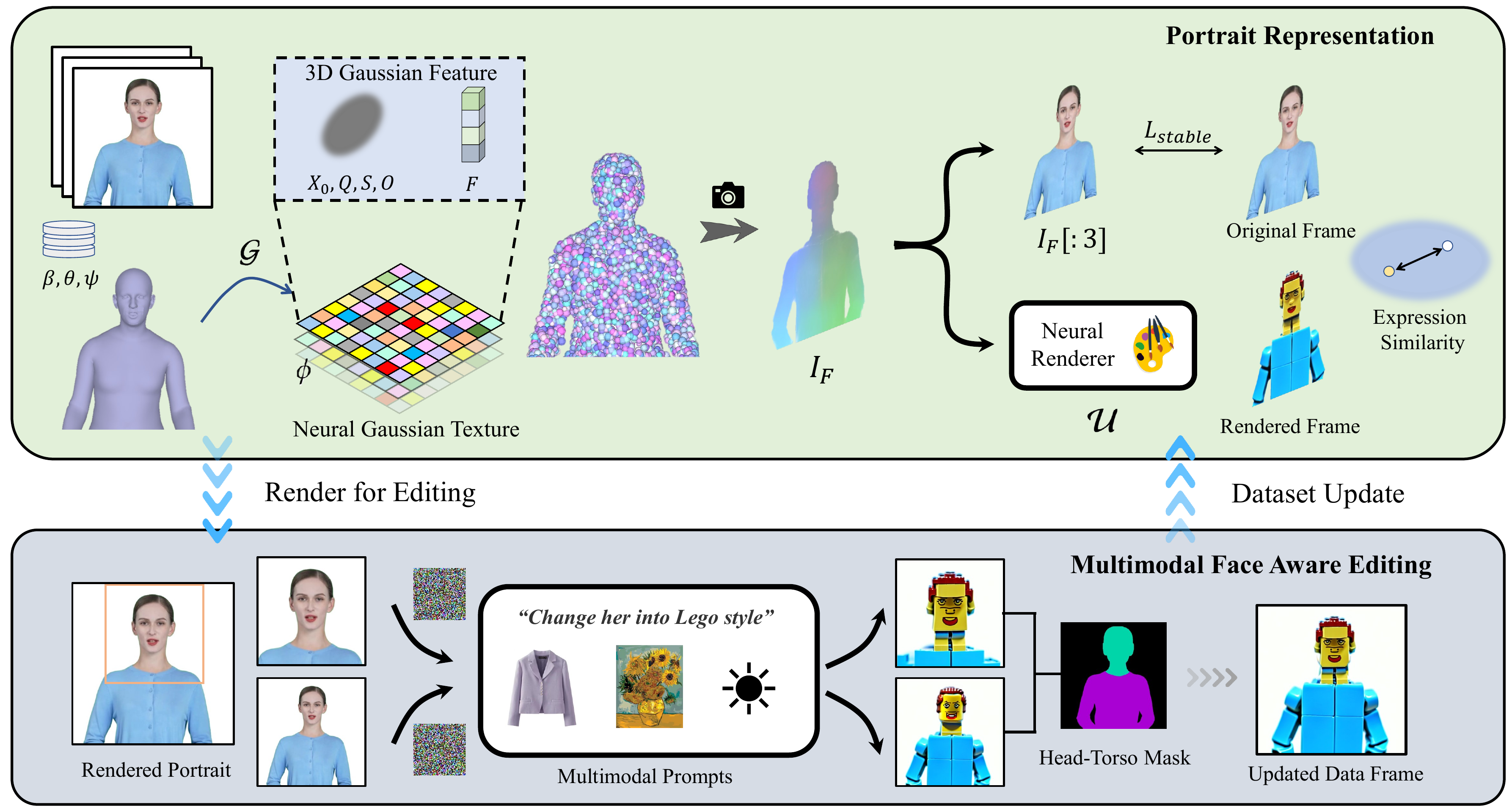}
  \caption{We first track the SMPL-X coefficients of the given monocular video, and then use a Neural Gaussian Texture mechanism to get a 3D Gaussian feature field. These neural Gaussians are further splatted to render portrait images. An iterative dataset update strategy is applied for portrait editing, and a Multimodal Face Aware Editing module is proposed to enhance expression quality and preserve personalized facial structures.}
  \label{fig:pipeline}
  \vspace{-2mm}
\end{figure*}

\section{Method}
As depicted in Fig.~\ref{fig:pipeline}, we develop a system that effectively distills knowledge from multimodal generative models to enable consistent, high-quality, and multimodal portrait video editing. To ensure consistency across frames, we propose a 3D portrait representation utilizing 3DGS and holistic human body priors (Sec.~\ref{sec:representation}). For high-quality rendering and expressive editing, we incorporate a Neural Gaussian Texture mechanism (Sec.~\ref{sec:gaussian_texture}). To support multimodal editing, we introduce specific techniques for text driven editing, image driven editing, and relighting. And we propose strategies to enhance the awareness of expressions and facial structures  (Sec.~\ref{sec:editing}). In the following, we first provide the preliminary knowledge of the 3DGS and SMPL-X models in Sec.~\ref{sec:preliminary}, and then introduce the technical details.

\subsection{Preliminary}
\label{sec:preliminary}
\subsubsection{3D Gaussian Splatting}

3DGS chooses 3D Gaussians as geometric primitives to represent scenes. Every Gaussian is defined by a 3D covariance matrix $\mathbf{\Sigma}$ centered at point $\mathbf{x_0}$:
\begin{equation}
  g(\mathbf{x}) = e^{-\frac{1}{2} (\mathbf{x}-\mathbf{{x_0}})^{T} \mathbf{\Sigma} ^{-1}(\mathbf{x}-\mathbf{x_0})}.
  \label{eq:gs}
\end{equation}
$\mathbf{\Sigma}$ is decomposed into a rotation matrix $R$ and a scaling matrix $\varLambda$ corresponding to learnable quaternion $\mathbf{q}$ and scaling vector $\mathbf{s}$:
\begin{equation}
  \mathbf{\Sigma} = R\varLambda\varLambda^TR^T.
  \label{eq:cov}
\end{equation}

Each 3D Gaussian is attached another two attributes: opacity $o$ and SH coefficients $\mathbf{h}$. The final color for a given pixel is calculated by sorting and blending the overlapped Gaussians:
\begin{equation}
  \mathbf{C} = \sum_{i\in N} \mathbf{c}_i\alpha_i\prod_{j=1}^{i-1} (1-\alpha_j),
  \label{eq:blend}
\end{equation}
where $\alpha_i$ is computed by the multiplication of projected Gaussian and $o$. Gaussian field can be denoted as  $\{\mathbf{x_0}, \mathbf{q}, \mathbf{s}, o, \mathbf{h}\}$.

\subsubsection{SMPL-X}

SMPL-X model~\cite{SMPL-X:2019} is a holistic, expressive body model, and is defined by a function $M\left(\beta, \theta, \psi \right):\mathbb{R}^{| \beta | \times | \theta | \times | \psi |} \rightarrow \mathbb{R}^{3V}$: 
\begin{align}
M   \left(\beta,\theta,\psi\right) &= W \left(T \left(\beta,\theta,\psi \right) , J\left(\beta \right),  \theta,\mathcal{W} \right), 									\label{skinning}			\\
T \left(\beta,\theta,\psi\right) &= \bar{T} + B_S \left(\beta;\mathcal{S} \right) + B_E  \left( \psi;\mathcal{E} \right) + B_P \left(\theta;\mathcal{P} \right).	\label{smpl_offsets}
\end{align}
  $\beta$, $\theta$, $\psi$ are shape, pose and expression parameters, respectively. $B_S \left(\beta;\mathcal{S} \right)$, $ B_P \left(\theta;\mathcal{P} \right)$, $ B_E \left(\psi;\mathcal{E} \right)$ are the blend shape functions. Blend skinning function $W(\cdot)$~\cite{lewis2000pose} rotates the vertices in $T\left(\cdot\right)$ around the estimated joints $J(\beta)$ smoothed by blend weights. To model long hairs and loose clothing, we introduce a learnable vertices displacement and the final mesh is computed as:
\begin{equation}
    \widehat{M}   \left(\beta,\theta,\psi\right) = M   \left(\beta,\theta,\psi\right) + \Delta M.
\end{equation}
\subsection{Portrait Representation}

To achieve high-fidelity and efficient rendering, we utilize dynamic 3DGS as the portrait avatar representation. Although naively using color or SH coefficient $\mathbf{h}$ may be enough for reconstruction tasks like previous representations~\cite{Zielonka2023Drivable3D,li2024animatablegaussians,xiang2024flashavatar}, it is not enough for editing task, especially for some complex styles. Many styles 
are not inherently 3D-consistent, as demonstrated in Fig.~\ref{fig:NGT_demonstrate}, directly fitting these signals with a 3D model may introduce blur or artifacts. Some styles also have complex structures, which is hard to be optimized for pure 3D models. To improve the representation ability and make it possible to edit with complex styles, we introduce a novel Neural Gaussian Texture mechanism.

\label{sec:representation}
\subsubsection{Neural Gaussian Texture}
\label{sec:gaussian_texture}

Similar to FlashAvatar~\cite{xiang2024flashavatar}, we maintain a 3D Gaussian field on the UV space of the SMPL-X model, and further deform the Gaussians according to the deformation of underlying meshes tracked from the input video. By embedding a 3D Gaussian field on the surface, the 3D Gaussian field could be efficiently transformed by parameters $\beta$, $\theta$, $\psi$. \revision{Inspired by Neural Texture proposed by Defered Neural Rendering~\cite{thies2019deferred}, we store learnable features for each Gaussian, instead of storing spherical harmonic coefficients. 
To be specific, we have a Neural Gaussian Field $\phi$ in the UV field where each pixel is characterized by four attributes: neural feature, opacity, scales, and rotation. Using UV mapping $\mathcal{G} \in \mathbb{R}^3 \rightarrow \mathbb{R}^2$, we transform neural Gaussians from UV space to 3D space.}
This operation $\mathcal{F}$ could be written as:
\begin{equation}
    (X_0, Q, S, O, F) = \mathcal{F}(\widehat{M}   \left(\beta,\theta,\psi\right) , \mathcal{G}, \phi),
\end{equation}
 Given $\widehat{M}   \left(\beta,\theta,\psi\right)$, $\mathcal{G}$ and $\phi$, we could get the embeded 3D Gaussian field $(X_0, Q, S, O, F)$ corresponding to a certain frame.

\subsubsection{Neural Rendering}
Given camera intrinsic parameters $K$, camera poses $P = \{P_i\}_{i=1}^{N}$, and the 3D Gaussian field, we perform differentiable tile renderer $\mathcal{R}$ to render a feature image. Then the feature image is operated by a 2D Neural Renderer $\mathcal{U}$ to convert it to RGB domain:
\begin{align}
    I_F &= \mathcal{R}(({X_0, Q, S, O, F }),K,P), \\
    I &= \mathcal{U}(I_F).
\end{align}
$I$ and $I_F$ share the same resolution, and they are all $512\times 512$ in our setting.  

\begin{figure}[t]
  \centering
  \includegraphics[width=\linewidth]{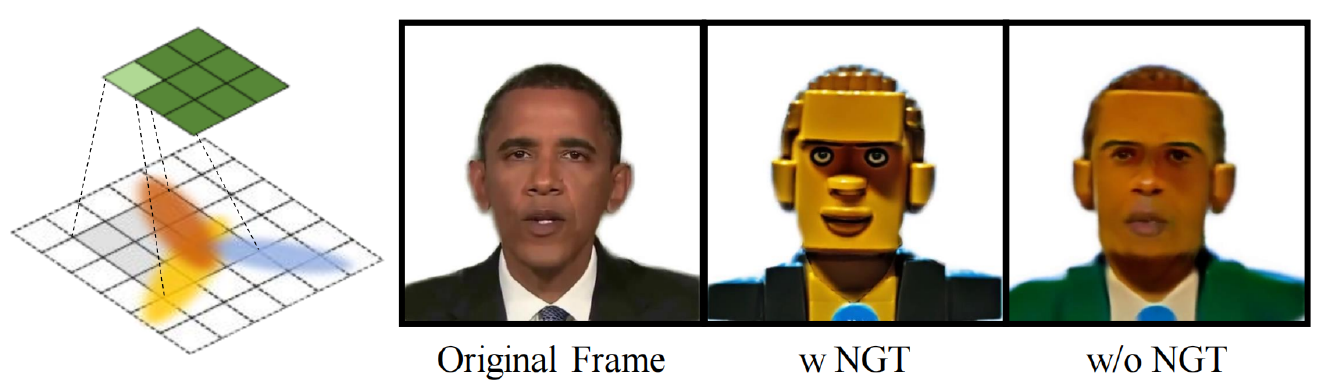}
  \caption{The Neural Renderer could effectively combine the information of splatted Gaussians and further improve the representation ability of 3D Gaussian portrait representation. With our Neural Gaussian Texture mechanism, the edited portrait follows prompts better and exhibit higher quality. (given instruction: Turn him into Lego style) }
  \label{fig:NGT_explain}
  \vspace*{-3mm}
\end{figure}

Many styles differ greatly from real people or are not totally 3D consistent. As shown in Fig. \ref{fig:NGT_explain}, our 2D Neural Renderer operates on the splatted feature map. Our Neural Gaussian Texture mechanism improves the model's capacity and could effectively combine the information of splatted Gaussians, which further improves the representation ability.


\subsubsection{Reconstruction Details}
We reconstruct the personalized 3D Gaussian Avatar with the following loss terms:

\paragraph{Reconstruction Loss.} This loss requires that the rendered result is consistent with the input RGB image, which is common for RGB reconstruction and can be formulated as:
\begin{equation}
L_{recon}(I,I_{src}) =  \|I - I_{src} \|_1.
\end{equation}

\paragraph{Mask Loss.} This loss requires that the rendered alpha channel $A$ is consistent with the segmentation map of the input source image:
\begin{equation}
L_{mask}(A,A_{src}) =  \|A - A_{src} \|_1.
\end{equation}

\paragraph{Perceptual Loss.}
The perceptual loss $L_{LPIPS}$ of \cite{zhang2018perceptual} is utilized to provide robustness to slight misalignments and shading variations and improve details in the reconstruction.
We choose VGG as the backbone of LPIPS.

\paragraph{Stable Loss.}
We found that training with above three loss terms may be unstable, and we further supervise part of the latent feature space $F$ directly:
\begin{equation}
L_{stable}(I_F,I_{src}) =  \|I_F[:3] - I_{src} \|_1,
\end{equation}
where the first 3 channels of $I_F$ are supervised by input source frames.

In summary, the overall loss of training our model is defined as:
\begin{equation}
\begin{split}
  L_{total} =& \lambda_{1}L_{recon}(I,I_{src})+\lambda_{2}L_{mask}(A,A_{src}) \\&+ \lambda_{3}L_{LPIPS}(I,I_{src}) + \lambda_{4}L_{stable}(I_F,I_{src}).
\end{split}
\end{equation}

\subsection{Editing}
\label{sec:editing}
\begin{figure}[t]
  \centering
  \includegraphics[width=\linewidth]{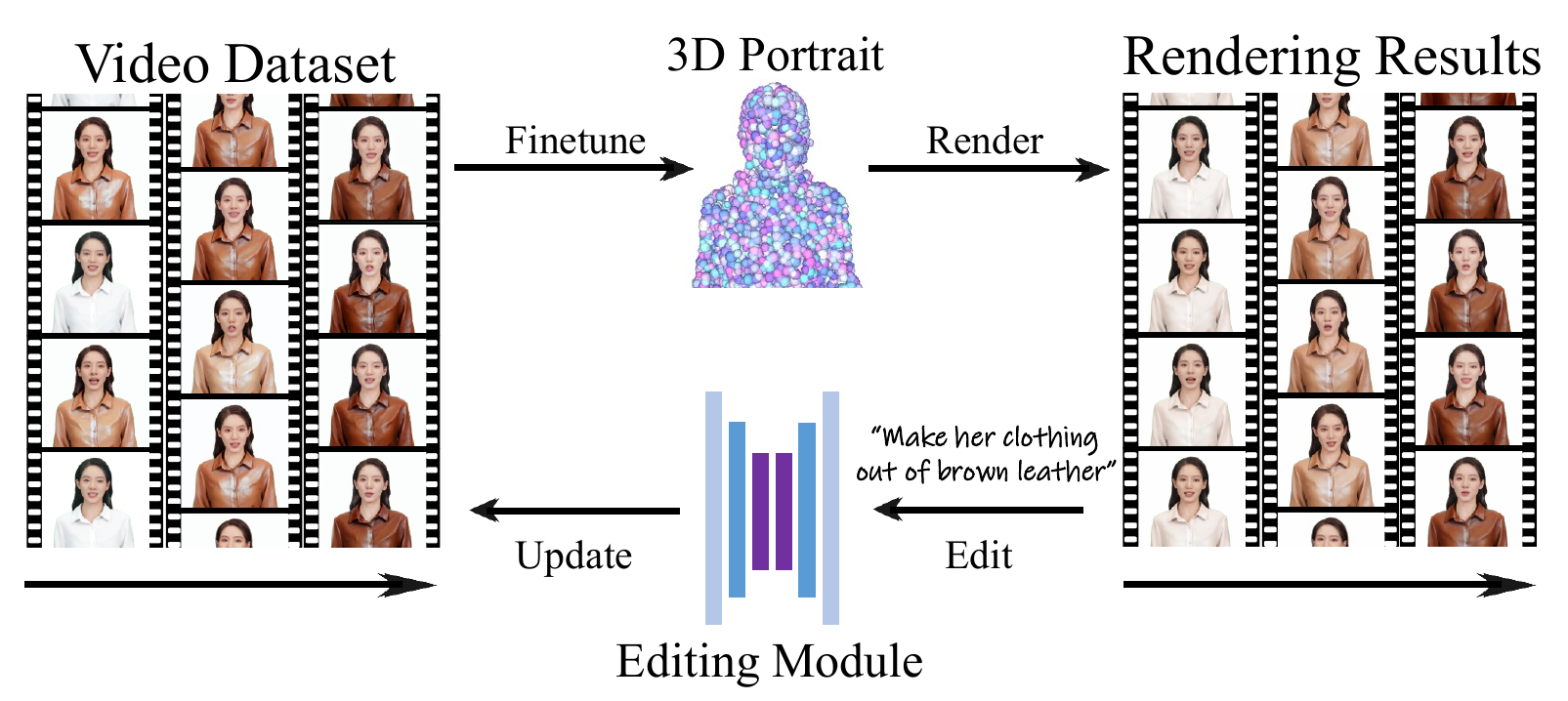}
  \caption{\revision{We alternate between editing the dataset of video frames and updating the underlying 3D portrait. The portrait model will gradually converge to the target prompt, achieving both 3D and temporal consistency.} }
  \label{fig:process}
  \vspace*{-3mm}
\end{figure}

We employ a variety of pre-trained generative models for multimodal prompt-guided editing. To tackle the issue of inconsistent edits across different frames, as illustrated in Fig. \ref{fig:process}, we alternate between editing the dataset of video frames and updating the underlying 3D portrait. \revision{Specifically, this process is to repeat as follows: (1) A portrait image is rendered from a training viewpoint. (2) The image is edited by the editing module. (3) The training dataset image is replaced with the edited image. (4) The portrait representation continues training with the updated dataset. The portrait model will gradually converge to the targeting prompt, achieving both 3D and temporal consistency.} 

To handle degradation problems in expressions and facial structures, we propose an expression similarity guidance term and a face-aware portrait editing module to emphasize facial information.

\begin{figure*}[t]
  \centering
  \includegraphics[width=\linewidth]{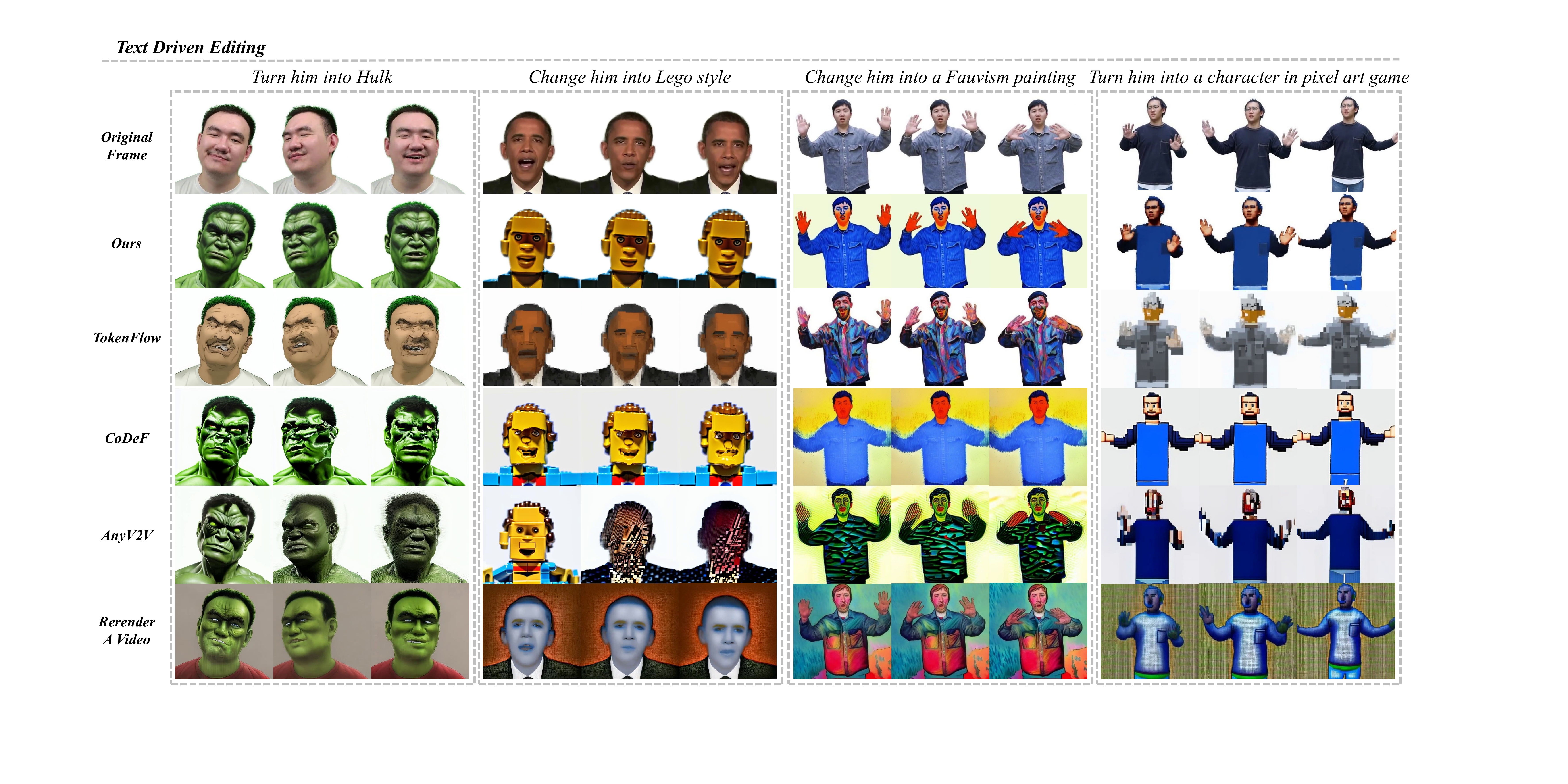}
  \caption{Qualitative comparisons on text driven portrait editing. }
  \label{fig:text_editing}
\end{figure*}

\subsubsection{Expression Similarity Guidance}
\label{part:expression}
Although many 2D editing models are claimed to be structure-preserving, they are not very robust to complex expression details. Accumulated errors after many times of editing may further misguide the expressions far from the original video. To enhance expression cognition, we map the rendered image and input source image to the latent expression space of EMOCA~\cite{filntisis2022visual}, and use a loss function to ensure similarity:
\begin{equation}
L_{exp}(I,I_{src}) =  \|\mathcal{E}_{exp}(I) - \mathcal{E}_{exp}(I_{src}) \|^2_2.
\end{equation}

\subsubsection{Face-Aware Portrait Editing}

When editing an upper body image where the face occupies a relatively small portion, the editing may not be robust enough to detailed facial structure. We further propose a training-free strategy that improves the editing quality of the face region. As shown in Fig.~\ref{fig:pipeline}, we first crop and resize the face region into $512\times512$. Both facial part and portrait part are then edited by the image editing model, and both edited parts are then composited into the final frame image with the head-torso mask.

\subsubsection{Editing Details}
For each optimization step, we randomly select a frame and use the corresponding SMPL-X parameters to render image $I$. The selected updated dataset frame is denoted as $I^*$. The corresponding original image (which is the image that is not edited) is denoted as $I_{src}$. We finetune the avatar reconstructed in section~\ref{sec:representation} with the following loss function:
\begin{equation}
\begin{split}
  L_{edit} = &\lambda_{1}L_{recon}(I,I^*)+\lambda_{2}L_{mask}(A,A_{src})  \\& + \lambda_{3}L_{LPIPS}(I,I^*)+\lambda_{4}L_{stable}(I_F,I_{src})+\lambda_{5}L_{exp}(I,I_{src}).
\end{split}
\end{equation}


\subsection{Applications}
 Our scheme is a unified portrait video editing framework. Any structure-preserving image editing model could be used to synthesize a 3D consistent and temporally coherent portrait video. In this paper, we demonstrate its effectiveness via several challenging tasks:

\subsubsection{Text Driven Editing}
\label{part:text}

We use InstructPix2Pix~\cite{brooks2022instructpix2pix} as a 2D editing model. We add partial noise to the rendered image and edit it based on input source image $I_{src}$ and instruction.


\subsubsection{Image Driven Editing}
We focus on two kinds of editing works based on image prompts. One kind is to extract the global style of a reference image and another aims to customize an image by placing an object at a specific location. These approaches are utilized in our experiments for style transfer and virtual try-on. We use the method of~\cite{gatys2016neural} to transfer the style of a reference image to the dataset frames and use AnyDoor~\cite{chen2023anydoor} to change the clothes of the subject.

\subsubsection{Relighting}
We utilize IC-Light~\cite{iclight} to manipulate the illumination of the video frames. Given a text description as the light condition, our method can harmoniously adjust the lighting of the portrait video.

%% file: Sec04_Experiments.tex
\section{Experiments}

\subsection{Implementation Details}
\revision{We use the videos released by NeRFBlendshape~\cite{Gao2022nerfblendshape}, Neural Head Avatar~\cite{grassal2022neural}, INSTA~\cite{zielonka2023instant} and PointAvatar~\cite{zheng2023pointavatar} for validation}. Since the released videos only contain the head region, we also collected some datasets from the Internet and captured some monocular videos of the upper body. \revision{We use FaRL~\cite{Zheng_2022_CVPR} to get head-torso masks.} We use an algorithm similar to TalkSHOW~\cite{yi2022generating} for fitting SMPL-X parameters to video frames. It takes about 10 minutes for reconstruction and about 20 minutes for editing. We run our experiments on one RTX 3090 GPU.

\begin{figure*}[t]
  \centering
  \includegraphics[width=\linewidth]{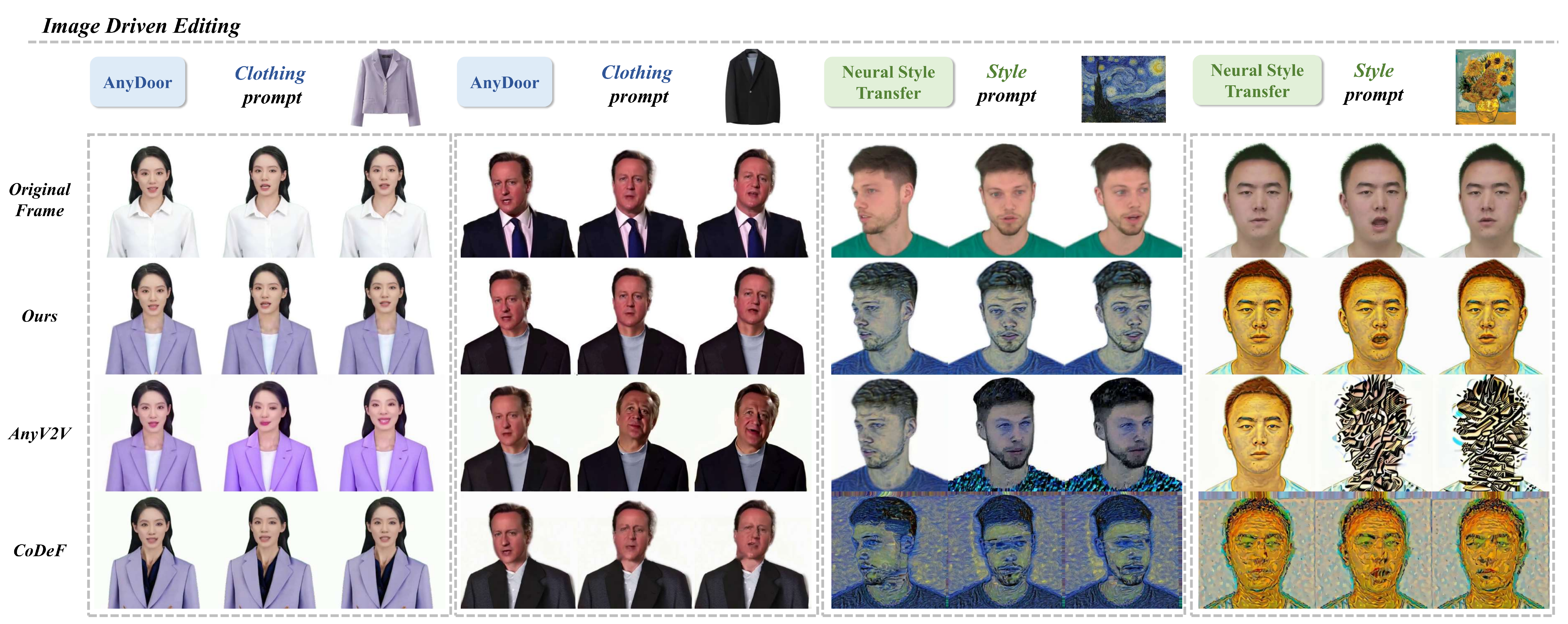}
  \caption{Qualitative comparisons on image driven portrait editing. }
  \label{fig:image_driven_editing}
\end{figure*}

\begin{figure*}[t]
  \centering
  \includegraphics[width=\linewidth]{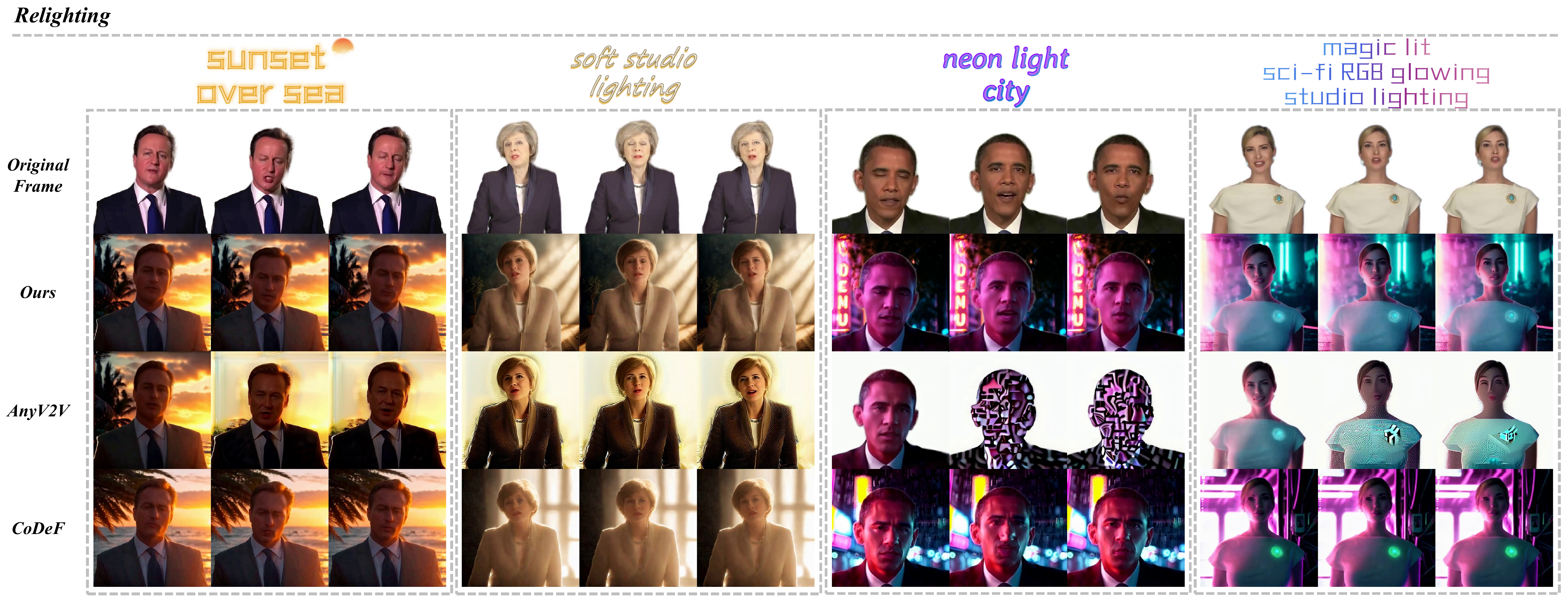}
  \caption{Qualitative comparisons on relighting.}
  \label{fig:relight}
\end{figure*}

\begin{figure*}[t]
  \centering
  \includegraphics[width=\linewidth]{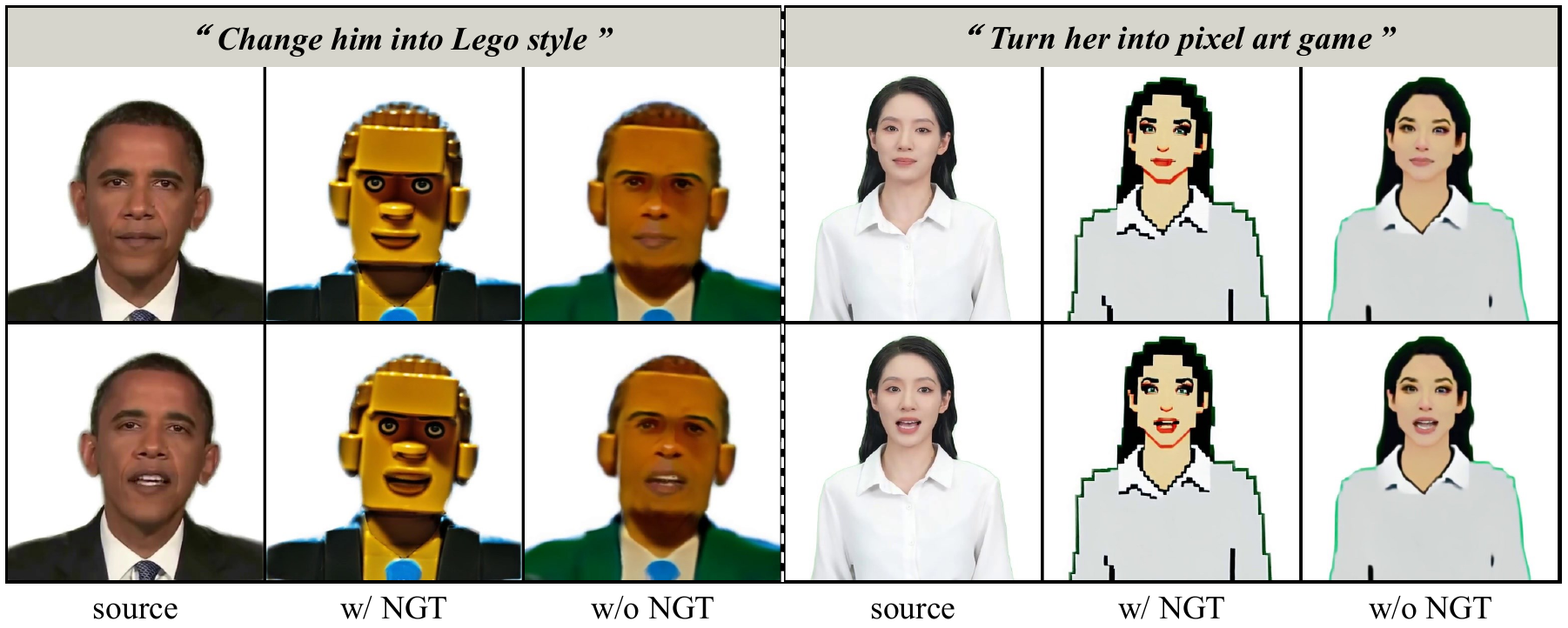}
  \caption{Neural Gaussian Texture mechanism could remarkably improve the editing results and make it possible to edit with more complex styles.}
  \label{fig:ablation_ngt}
\end{figure*}
\subsection{Qualitative Comparison}

We compare our method with state-of-the-art video editing methods, including TokenFlow~\cite{tokenflow2023}, Rerender-A-Video (denoted as RAV)~\cite{yang2023rerender}, CoDeF~\cite{ouyang2023codef} and AnyV2V~\cite{ku2024anyv2v}. TokenFlow and Rerender-A-Video only support text-driven editing tasks, while CoDeF and AnyV2V could support editing with all modalities. For CoDeF, we train the deformation field and canonical image on the input video first. Then, we edit the canonical image and generate the final edited video according to the deformation field. For AnyV2V, we edit the first frame and then perform image-to-video reconstruction.

To ensure a fair comparison, we limit the video segments used in our evaluation to 2 seconds, each consisting of 60 frames. This is necessary because TokenFlow requires significant GPU memory as the number of frames increases, and CoDeF must learn the deformation fields for the entire video sequence, making it unsuitable for long videos. Although our method can handle videos of arbitrary length, selecting shorter segments allows for a fair evaluation across different methods.

 We present qualitative comparisons on text-driven editing in Fig.~\ref{fig:text_editing}, image-driven editing in Fig.~\ref{fig:image_driven_editing}, and relighting in Fig.~\ref{fig:relight}. For TokenFlow and Rerender-A-Video, we observe that sometimes the expressions in the edited frames do not maintain consistency with the original video, and the edits fail to align with the given prompts. This may be because extended attention mechanisms can cause the latent codes to drift out of the domain, thereby degrading the quality of the edited results. Additionally, both methods frequently produce noticeable artifacts in the facial regions. This discrepancy can be attributed to the limitations of extended attention in maintaining detailed consistency, especially in capturing facial expressions. Inaccurate correspondences in nearest neighbor search or optical flow estimation further exacerbate these discrepancies. 

Although CoDeF's unique modeling approach enhances its capacity to preserve detailed consistency in short video segments, it fails to generate reasonable results when faced with exaggerated expression and pose changes. This issue primarily stems from the limitations of its 2D deformation field, which is inadequate for modeling complex 3D portrait deformations. We also observe that AnyV2V lacks stability in portrait editing. It frequently fails to maintain consistent appearance and structural integrity, likely due to its unstable editing scheme. 

In contrast, our approach leverages a 3DGS-based portrait as the geometric representation, which ensures superior 3D consistency. By integrating prior information about the portrait, we precisely capture changes in expressions and postures, thereby maintaining temporal consistency in the edited results. Moreover, our model adeptly handles challenging multimodal prompts, which can be problematic for other methods. For a more detailed comparison, we encourage viewing the accompanying video.

\subsection{Quantitative Comparison}

\begin{table}[b]
\begin{tabular}{|c|c|c|c|c|c|}
\cline{1-6}
                           & TokenFlow  & CoDeF & AnyV2V & RAV & Ours  \\ \cline{1-6}
Q1 & 8.0 & 19.1 & 3.3 & 3.4 & \textbf{66.2} \\ \cline{1-6}
Q2 & 6.8 & 7.1 & 2.6 & 6.9 & \textbf{76.6} \\ \cline{1-6}
Q3 & 3.9 & 6.1 & 1.2 & 3.5 & \textbf{85.3} \\ \cline{1-6}
Q4 & 3.8 & 5.1 & 1.4 & 4.3 & \textbf{85.4} \\ \cline{1-6}
Q5 & 4.5 & 6.7 & 1.4 & 2.2 & \textbf{85.2} \\ \cline{1-6}
\end{tabular}
\caption{The table reports the percentages at which a method was rated the best with respect to a specific question. Our method remarkably outperforms other methods in all questions, which demonstrates that our approach is much more likely to be favored by users.}
\label{tab:userstudy}
\end{table}

We conducted a user study to further quantitatively validate our method. Participants were asked to watch rendered videos side by side from various methods and respond to a series of questions comparing the results. For each group of editing results, participants addressed the following queries: 
 \begin{itemize}
\item Q1: Which method best follows the given input prompt? (Prompt Preservation)
\item Q2: Which method best retains the identity of the input sequence in the video? (Identity Preservation)
\item Q3: Which method best maintains temporal consistency in the video? (Temporal Consistency)
\item Q4: Which method best preserves expressions and body movements of the input sequence in the video? (Human Motion Preservation)
\item Q5: Which method is best overall considering the above four aspects? (Overall)
\end{itemize}

\revision{We collected statistics from 96 participants across 23 groups of editing results.} For each case, the video results were randomly shuffled for fair comparison. As shown in Table.~\ref{tab:userstudy}, our method remarkably outperforms other methods in prompt preservation, identity preservation, temporal consistency, and human motion preservation and is rated as the best in overall quality. These results demonstrate that our approach is highly favored by users, highlighting its effectiveness in various editing dimensions.

\begin{figure*}[t]
  \centering
  \includegraphics[width=\linewidth]{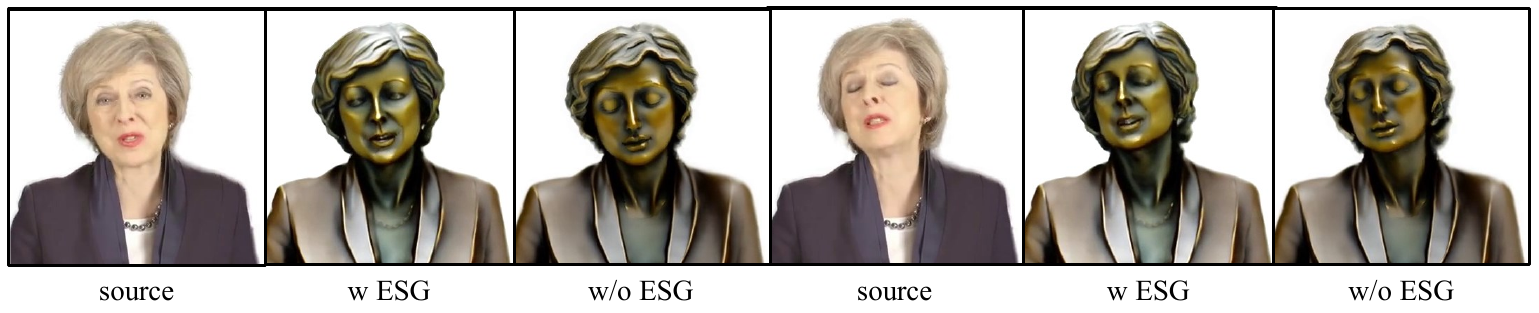}
 
  \caption{Expression Similarity Guidance could effectively solve expression degradation problems and keep the expressions consistent with original video frames. (prompt: Change her into a bronze statue.)}
  \label{fig:ablation_esg}
  \vspace*{-3mm}
\end{figure*}

\begin{figure}[t]
  \centering
  \includegraphics[width=0.93\linewidth]{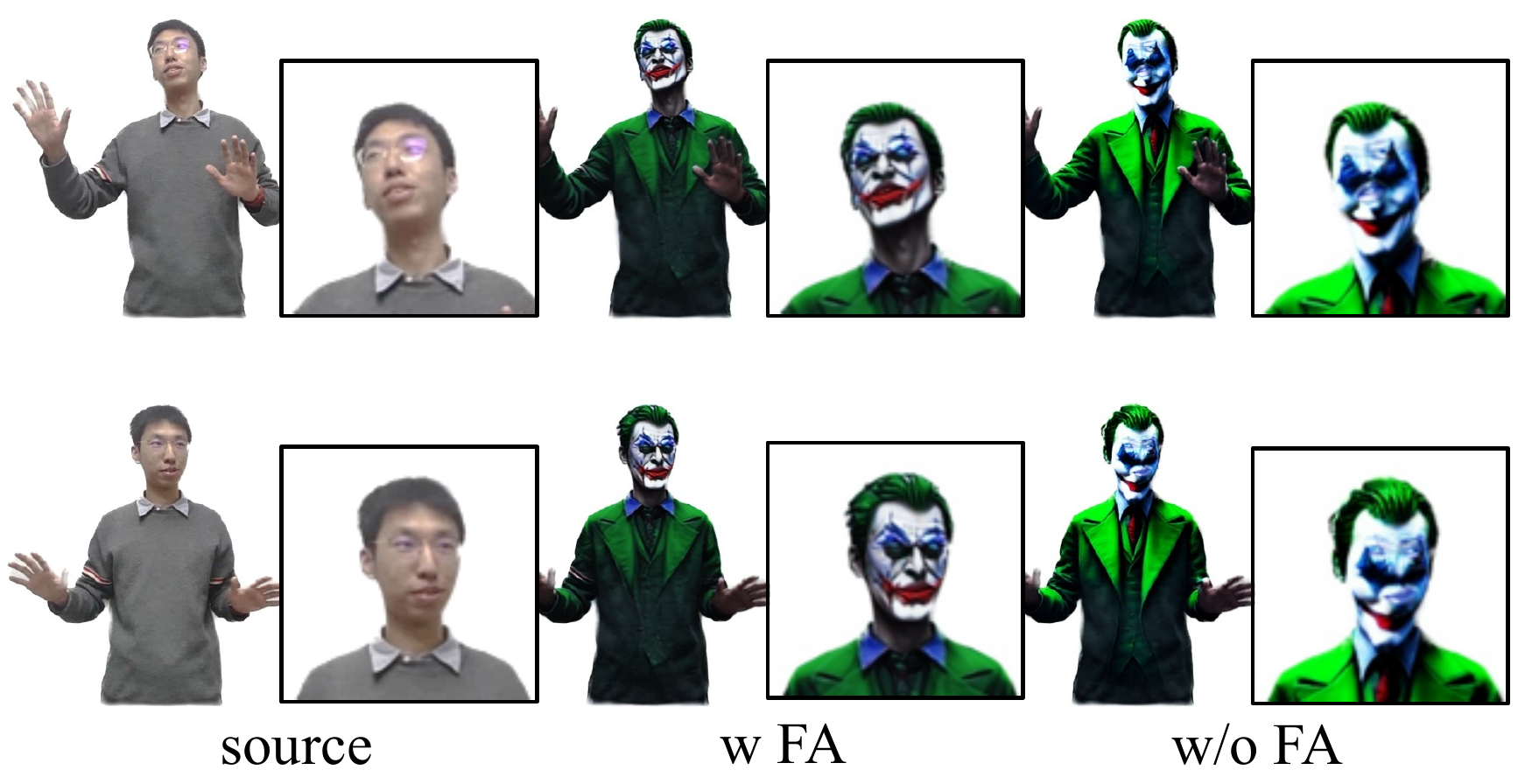}
  
  \caption{Naively editing the whole portrait image may cause misalignment of head pose, and blur in the facial region.}
  \label{fig:ablation_fa}
  \vspace*{-4mm}
\end{figure}

\subsection{Editing Efficiency}

We further validated the editing efficiency of our method by analyzing the number of frames processed per minute, as illustrated in Table.~\ref{tab:efficiency}. Different methods employ different modes of inference. For a fair comparison, we compute the time cost including both reconstruction and editing, and average it across the processed frames. We can see that our method outperforms previous video editing methods in terms of efficiency, which further showcases its promising application prospects.

\begin{table}[h]

\begin{tabular}{|c|c|c|c|c|}
\cline{1-5}
TokenFlow  & CoDeF & AnyV2V & RAV & Ours  \\ \cline{1-5}
1.6 & 5.0 & 1.6 & 10.0 & \textbf{60.0} \\ \cline{1-5}
\end{tabular}
\caption{Comparison in editing efficiency. The values represent the number of frames edited per minute. Our method outperforms previous video editing methods, further verifying its promising application prospects.}
\label{tab:efficiency}
\vspace{-8mm}
\end{table}

%% file: Sec05_Ablation.tex
\section{Ablation Study}
\subsection{Neural Gaussian Texture}
\label{abl:NGT}

Previous portrait representations~\cite{xiang2024flashavatar,qian2023gaussianavatars} use explicit 3D Gaussian for rendering by storing spherical harmonic coefficients for each Gaussian to directly render portrait image. We demonstrate that this approach is unable to represent complex styles like contour lines and brush strokes as they adopt pure 3D representations.

Fig.~\ref{fig:ablation_ngt} shows the comparison results between using our Neural Gaussian Texture (NGT) and explicit 3D Gaussian. For the prompt ``Change him into Lego style'', our method adeptly transforms the editing into the desired shape, while explicit 3D Gaussians struggle to achieve a Lego-like deformation. This is because our Neural Renderer could fuse the features of splatted Gaussians, and further improve its representation ability. For the prompt ``Turn her into pixel art game'', explicit 3D Gaussians fail to represent contour lines and pixel style elements, demonstrating the limitations of using a purely 3D consistent representation for such stylized edits.

\subsection{Face-Aware Portrait Editing}

When editing an upper body image where the face occupies a relatively small portion, the model's editing may not be robust enough to head pose and facial structure. Face-Aware Portrait Editing (FA) could enhance the awareness of face structures by performing editing twice. As demonstrated in Fig.~\ref{fig:ablation_fa}, naively editing the whole portrait image may cause misalignment of head pose, and blur in the facial region.

\subsection{Expression Similarity Guidance}

By mapping the rendered image and input source image into the latent expression space of EMOCA, and optimizing for expression similarity, we can further keep the expressions natural and consistent with the original input video frames. As demonstrated in Fig.~\ref{fig:ablation_esg}, omitting Expression Similarity Guidance during training leads to expression degeneration.

%% file: Sec06_Conclusion.tex
\section{Conclusion \& Discussions}

We proposed an expressive multimodal portrait video editing scheme. In contrast to previous approaches that primarily focus on the 2D domain, we elevated the portrait video editing challenge to a 3D perspective. Our method embedded a 3D Gaussian field onto the surface of SMPL-X, ensuring consistency in human body structures across both spatial and temporal domains. Additionally, the proposed Neural Gaussian Texture mechanism could effectively deal with complex styles and achieve rendering speeds of over 100FPS. We leveraged the multimodal editing knowledge of 2D generative models to enhance the quality of 3D editing. Our expression similarity guidance and face-aware portrait editing module effectively handled the degradation problems of iterative dataset updates.

Although we have achieved remarkable improvement in quality and efficiency compared with existing works, there still remain some limitations. Our method relies on tracked SMPL-X, and thus large errors in tracking may cause artifacts. As our method utilizes pre-trained 2D editing models for dataset update, the editing ability of our method is restricted by these models. We believe more powerful 2D editing models will further unleash the potential of our paradigm.
\begin{acks}
This research was supported by the National Natural Science Foundation of China (No.62122071, No.62272433, No.62402468), the Fundamental Research Funds for the Central Universities (No. WK3470000021), and the advanced computing resources provided by the Supercomputing Center of University of Science and Technology of China.
\end{acks}